\documentclass{article}
\usepackage{hyperref}
\usepackage{spconf,amsmath,graphicx}
\usepackage[protrusion=true,expansion=true,stretch=20,shrink=20]{microtype}
\usepackage{cite}
\usepackage{amssymb,amsfonts}
\usepackage{textcomp}
\usepackage{xcolor}
\usepackage{booktabs}
\usepackage{hyperref}
\usepackage{tikz}
\usetikzlibrary{patterns,patterns.meta,shapes.geometric,arrows.meta,positioning,calc}
\usepackage{pgfplots}
\pgfplotsset{compat=1.18}
\let\oldthebibliography\thebibliography
\renewcommand{\thebibliography}[1]{\oldthebibliography{#1}\setlength{\itemsep}{0pt}\setlength{\parsep}{0pt}}

\title{AVCC: Audio-Visual Conversation Corpus for Turn-Taking Prediction with a Human Benchmark}
\name{Haotian Qi and Gabriel Skantze}
\address{Department of Speech, Music and Hearing,
         KTH Royal Institute of Technology, Stockholm}

\begin{document}
\ninept
\maketitle
\begin{abstract}
Predicting who will speak next, and when, is a prerequisite for conversational
agents and social robots that take turns without waiting for silence.
Multiparty next-speaker prediction (NSP) has been studied on corpora with
per-speaker sensors, each with its own event definitions. No public benchmark
covers the single camera and microphone a robot has, and no model has been
compared against how well people perform the same prediction. We turn AVCC, a
31-hour audio-visual corpus of two- and three-speaker online conversation
recorded with a single static camera and microphone, into a public NSP
benchmark. We add a speaker-disjoint seven-video test partition with 865 hold
and 865 shift events at mutual-silence gaps, word-level forced-aligned
transcripts for the whole corpus, and a human reference benchmark collected
at the same events. We score an audio-only Role-Relative VAP model (RoleVAP)
and the multimodal MuVAP model. Labels, transcripts, per-response human data,
and the scripts that reproduce every number are released.
\end{abstract}
% ===============================================================
\section{Introduction}
\label{sec:intro}

Speaker transitions in conversation take about 200\,ms on
average~\cite{stivers2009universals}, less than the time needed to plan an
utterance, so listeners must detect turn boundaries and even project the end of the current turn before it
arrives~\cite{levinson_torreira:15,de2006projecting}. They draw on prosody,
syntax and lexical content, gaze, and mouth and head
movement~\cite{duncan1972some,Duncan1977FacetofaceIR,kendon1967gaze,gravano2011turn}.
In groups of three or more, the same moment holds a second decision: who among
several candidates will take the floor~\cite{sacks:74}. Conversational agents
and social robots that take turns at a human pace face both decisions and
cannot wait for silence to resolve them~\cite{Skantze2021}. We refer to the
combined task as \textit{next-speaker prediction} (NSP), and to the coarser question of
whether the current speaker will continue or yield as the \emph{hold}/\emph{shift}
decision.

Computational models of turn taking have followed two lines. One predicts the
next speaker in multiparty conversation from gaze, head pose, mouth movement,
and respiration~\cite{jokinen2013gaze,ishii2016mouth,
ishii2016respiration,malik2020speaks,bohus2011decisions}
, typically on meeting or laboratory corpora recorded
with eye trackers, close-talk microphones, or a camera per participant, and
each study defines and evaluates its own set of turn events. The other treats
turn taking as continuous prediction from the speech signal, from end-of-turn
detection at silences through acoustic and lexical
models~\cite{roddy2018,ekstedt2020turngpt} to Voice Activity Projection
(VAP)~\cite{vap}, which predicts the future voice activity of every speaker at
every frame; this line is trained and evaluated on dyadic telephone corpora such
as Switchboard~\cite{godfrey1992switchboard}, Fisher~\cite{cieri2004fisher},
and Switchboard-NXT~\cite{calhoun2010nxt}, with one clean channel per speaker.
The lines have begun to meet: VAP has been extended to three
speakers~\cite{elmers2025triadic}, combined with gaze from head-worn
sensors~\cite{heo2025gaze}, and reformulated to map any number of speakers onto
a current and a next floor holder so that one model applies to groups of any
size~\cite{muvap}. Neither line has a public benchmark for the setting most social
robots work in: one camera and one microphone for the whole group. Nor is
there a shared set of prediction points on which systems can be compared, or
a measure of how well people perform the same prediction. Without that
measure we cannot tell whether a model at 65\% accuracy is far from or close
to what the signal allows.

\begin{table}[t]
\centering
\caption{Corpora relevant to multiparty NSP.
  Spk: speakers per recording. Sns.: sensors, S = one camera at most and one
  audio channel, M = per-speaker microphones or several cameras.
  Mod.: modality, A = audio, V = video. Trans.: transcripts. NSP: hold/shift
  or next-speaker labels. Hum.: human NSP reference.
  $\sim$ = partial.}
\label{tab:corpora}
\setlength{\tabcolsep}{1.6pt}
\footnotesize
\begin{tabular}{@{}lcccccccc@{}}
\toprule
& \multicolumn{4}{c}{\textbf{Corpus}} & \multicolumn{3}{c}{\textbf{Annotations}} & \\
\cmidrule(lr){2-5}\cmidrule(lr){6-8}
\textbf{Name} & \textbf{Spk} & \textbf{Hrs} & \textbf{Sns.} & \textbf{Mod.}
  & \textbf{Trans.} & \textbf{NSP} & \textbf{Hum.} & \textbf{Year} \\
\midrule
Switchboard~\cite{godfrey1992switchboard}  & 2    & 260   & M  & A   & \checkmark & -- & -- & 1992 \\
Fisher~\cite{cieri2004fisher}              & 2    & 2000  & M  & A   & \checkmark & -- & -- & 2004 \\
AMI~\cite{carletta2006ami}                 & 3--5 & 100   & M  & A+V & \checkmark & -- & -- & 2006 \\
NoXi~\cite{cafaro2017noxi}                 & 2    & 27    & M  & A+V & \checkmark & -- & -- & 2017 \\
AVA-AS~\cite{roth2020ava}                  & 2+   & 38    & S  & A+V & -- & -- & -- & 2020 \\
MultiMediate~\cite{mueller2021multimediates} & 3--4 & 16  & M  & A+V & -- & $\sim$ & -- & 2021 \\
MSDWild~\cite{msdwild}                     & 2+   & 32    & S  & A+V & -- & -- & -- & 2022 \\
CANDOR~\cite{reece2023candor}              & 2    & 850   & M  & A+V & \checkmark & -- & -- & 2023 \\
EgoCom~\cite{northcutt2023egocoms}         & 3    & 38.5  & M  & A+V & \checkmark & $\sim$ & -- & 2023 \\
WASD~\cite{roxo2024wasd}                   & 2+   & 30    & S  & A+V & -- & -- & -- & 2024 \\
TurnBench~\cite{jiang2026turnbench}        & 2    & 30    & S  & A   & \checkmark & \checkmark & -- & 2026 \\
\midrule
\textbf{AVCC (ours)} & \textbf{2--3} & \textbf{31} & \textbf{S} & \textbf{A+V} & \checkmark & \checkmark & \checkmark & \textbf{2026} \\
\bottomrule
\end{tabular}
  \vspace{-1.5ex} 
\end{table}

\begin{figure}[t]
\centering
\includegraphics[width=\columnwidth]{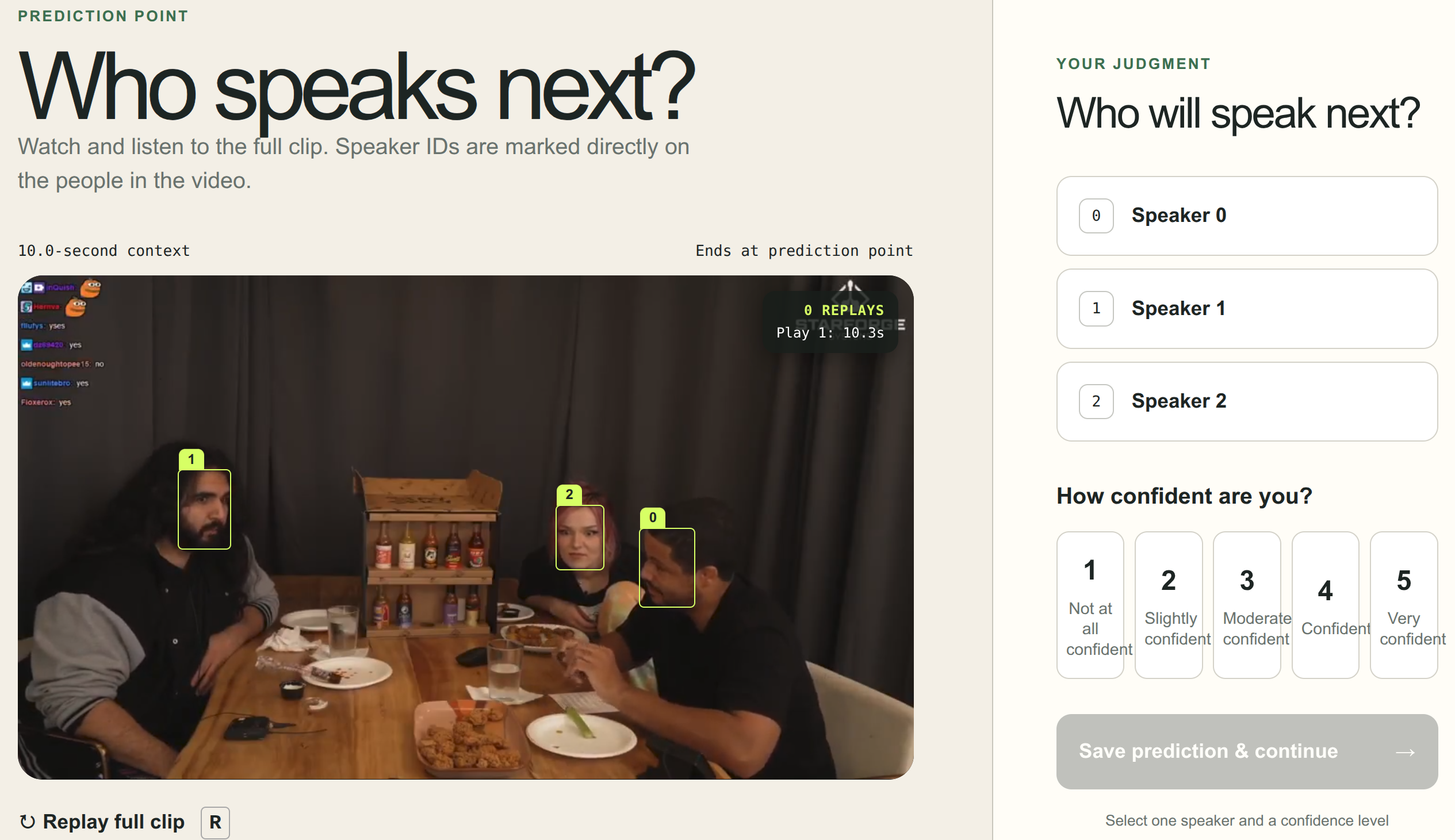}
\caption{The study interface. Participants watch a clip of up to 10\,s with
face labels (\textbf{0}, \textbf{1}, \textbf{2} for three-speaker clips)
drawn on every frame, then choose the next speaker and rate their confidence.}
\label{fig:interface}
  \vspace{-1.5ex} 
\end{figure}

Existing audio-visual corpora do not close the gap (Table~\ref{tab:corpora}).
AMI~\cite{carletta2006ami} records meetings with close-talk microphones and a
camera array; EgoCom~\cite{northcutt2023egocoms} uses one egocentric camera per
participant and provides segment-level speaking labels rather than events at
gap boundaries; CANDOR~\cite{reece2023candor} records dyadic video calls with
one stream per participant. NoXi~\cite{cafaro2017noxi},
AVA-ActiveSpeaker (AVA-AS)~\cite{roth2020ava}, MSDWild~\cite{msdwild},
WASD~\cite{roxo2024wasd}, and MultiMediate~\cite{mueller2021multimediates}
target active-speaker detection or group behavior and contain neither
hold/shift events nor a human NSP reference. TurnBench~\cite{jiang2026turnbench}
evaluates end-of-turn detection across six audio-only dyadic domains, with no
video, no multiparty data, and no human reference. None of these combines the
single camera and single microphone that most social robots have with
multiparty data, transcripts for models that use lexical cues, event labels at
prediction points, and a human reference measured at those same points.

AVCC\footnote{\url{https://github.com/Haotian-Qi/AVCC}}, the corpus used here, was introduced in~\cite{muvap} together with
MuVAP: 31 hours of unscripted two- and three-speaker conversation from 29
online videos, each recorded without cuts from one static camera and one
monaural track, released with hand-corrected voice activity labels and face
tracks. That release was designed for training: it has a training and a
validation partition but no held-out test data, its hold/shift events served
to fit and validate a probe on the validation partition rather than to compare
systems, it has no transcripts, and there is no measure of how well people
perform the same task, so a new model can be trained on AVCC but not compared with other systems or human observers. This paper turns AVCC into a
benchmark. Following~\cite{muvap}, a \emph{turn event} is a mutual-silence gap
between two segments of speech, labeled \emph{shift} when the floor changes
speaker and \emph{hold} when the same speaker continues. We contribute:

\begin{itemize}
  \item \textbf{Test partition and transcripts}: a speaker-disjoint
        seven-video test partition with 865 hold and 865 shift events,
        word-level forced-aligned transcripts for the whole corpus, and the
        scripts that produce them and every number in this paper.
  \item \textbf{Human benchmark}: a study with 154 participants
        (Fig.~\ref{fig:interface}) who predicted the next speaker from up to
        10\,s of audio-visual context, with a confidence rating per
        prediction, at exactly the prediction points used for models.
  \item \textbf{Baselines and analysis}: an audio-only RoleVAP model~\cite{vap,muvap}, and the multimodal MuVAP
        model~\cite{muvap}, scored on the same events under next-speaker
        identity and hold/shift scoring and compared with the participants by
        speaker count and by confidence.
\end{itemize}

% ===============================================================
\section{AVCC: Corpus and Annotations}
\label{sec:annotations}

\subsection{Corpus Overview}

\begin{figure}[t]
\centering
\begin{tikzpicture}
\begin{axis}[name=top, width=\columnwidth, height=4cm, ybar, bar width=4.5pt,
  ylabel={Duration (h)}, xlabel={Videos sorted by duration},
  xtick=\empty, xmin=-0.8, xmax=28.8, ymin=0, ymax=2.3, ytick={0,0.5,1,1.5,2},
  unbounded coords=discard, legend columns=4,
  legend style={font=\tiny, column sep=2pt},
  ]
\addplot[bar shift=0pt, fill=blue!35, draw=none] table[x=idx, y=d2train] {dataset/fig_videos.dat};
\addlegendentry{2-spk train/val}
\addplot[bar shift=0pt, fill=blue!80!black, draw=none] table[x=idx, y=d2test] {dataset/fig_videos.dat};
\addlegendentry{2-spk test}
\addplot[bar shift=0pt, fill=orange!45, draw=none] table[x=idx, y=d3train] {dataset/fig_videos.dat};
\addlegendentry{3-spk train/val}
\addplot[bar shift=0pt, fill=orange!95!black, draw=none] table[x=idx, y=d3test] {dataset/fig_videos.dat};
\addlegendentry{3-spk test}
\end{axis}
\end{tikzpicture}
\caption{Duration of each of the 29 videos by speaker count
and partition (longest 1\,h\,55\,min, shortest 20\,min, median 59\,min).}
\label{fig:corpus}
  \vspace{-1.5ex} 
\end{figure}
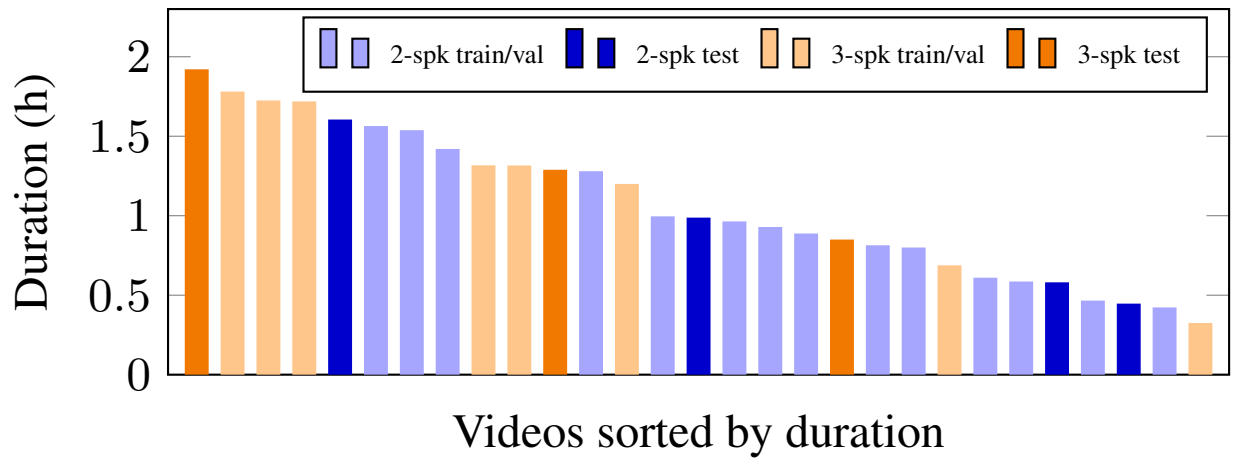

AVCC contains 31 hours of unscripted conversation from 29 publicly available
online videos (Table~\ref{tab:avcc_tab}, Fig.~\ref{fig:corpus}). Videos were
selected for continuous, unedited recording from one static camera with all
speakers normally in view, and one monaural audio track; there are no cuts,
no per-speaker microphones, and no camera motion. Long videos are stored as
several contiguous files (segments); an event's context window never crosses
a segment boundary. The videos are podcast
interviews, roundtable discussions, and gaming commentary. We count as
speakers the people who produce at least 5\% of the speech in a video; by
this rule 18 videos have two speakers and 11 have three (one video includes
a third person who speaks for about 1\% of the time; nine videos also contain a
bystander, labelled B, who is heard but not seen and is excluded from all counts). Partitions are assigned at the video level, so no speaker in the
test partition appears in training or validation. The seven test videos,
four with two speakers and three with three, cover all three genres and hold
417 hold and 417 shift events from two-speaker videos and 448 of each from
three-speaker videos. Events are unevenly distributed over videos: one
two-speaker video contributes 283 of the 417 two-speaker pairs and one
three-speaker video 229 of the 448 three-speaker pairs.

From the corrected voice activity labels, the floor changes 14.3 times per
minute in two-speaker videos and 16.4 times per minute in three-speaker
videos. The most talkative speaker
produces a median of 58\% of the speech in two-speaker videos and 50\% in
three-speaker videos. These statistics, the durations in
Table~\ref{tab:avcc_tab}, and Fig.~\ref{fig:corpus} are produced by a released
script from the annotation files.

\begin{table}[t]
\centering
\caption{AVCC corpus overview. Durations are measured from the video files
and rounded to the minute.}
\label{tab:avcc_tab}
\small
\begin{tabular}{@{}lcc@{}}
\toprule
\textbf{Subset} & \textbf{Videos} & \textbf{Duration} \\
\midrule
2-speaker      & 18     & 16h52m\\
3-speaker      & 11     & 14h07m\\
\midrule
Training       & 17    & 20h39m\\
Validation     & 5      & 2h41m\\
Test           & 7      & 7h41m\\
\textbf{Total} & \textbf{29} & 31h00m\\
\bottomrule
\end{tabular}
  \vspace{-1.5ex} 
\end{table}

\subsection{Annotation Pipeline}
\label{sec:pipeline}

\textbf{Face tracking.} Faces are detected and identified per frame with
InsightFace~\cite{deng2019arcface,guo2022scrfd}: SCRFD
detection, ArcFace identity embeddings, and Hungarian
assignment~\cite{kuhn1955hungarian} to per-video speaker identities. All
tracks were checked by hand for identity swaps between overlapping faces,
tracks lost across segment boundaries, and detections on photographs.

\textbf{Voice activity.} Speech labels were initialized with
TalkNet~\cite{talknet} at its default confidence threshold, after which every segment was corrected with our own annotation tool.

\textbf{Transcripts and alignment.} Each speaker turn was transcribed with
Voxtral~\cite{voxstral}. We inspected all utterance-level transcripts by hand,
marking non-English speech as \texttt{[foreign]} and 
unintelligible speech as \texttt{[unk]}; laughter is not transcribed.
Words were then force-aligned with
the Montreal Forced Aligner~\cite{mcauliffe2017montreal} (English acoustic
model, CMU dictionary), constrained to the transcribed segment, then all errors were corrected by hand.

\begin{figure}[t]
\centering
\begin{tikzpicture}[x=0.52cm, y=0.62cm, font=\small]

  %% --- background: context (light gray) and withheld (hatched) ---
  \fill[gray!15] (0,0.2) rectangle (10,4.4);
  \fill[pattern=north east lines, pattern color=gray!40]
    (10,0.2) rectangle (13,4.4);

  %% === Spk A (top row) ===
  \fill[gray!55] (0,3.20)    rectangle (3.5,3.75);
  \fill[gray!55] (5.0,3.20)  rectangle (8.5,3.75);
  \fill[gray!55] (11.2,3.20) rectangle (13,3.75);
  \draw[black!40] (0,3.20) rectangle (13,3.75);
  \node[anchor=east, font=\small\bfseries] at (-0.2,3.475) {Spk\,A};

  %% === Spk B (middle row) ===
  \fill[gray!55] (3.5,2.20) rectangle (5.0,2.75);
  \draw[black!40] (0,2.20) rectangle (13,2.75);
  \node[anchor=east, font=\small\bfseries] at (-0.2,2.475) {Spk\,B};

  %% === Spk C (bottom row) ===
  \fill[gray!55] (1.0,1.20) rectangle (2.2,1.75);
  \draw[black!40] (0,1.20) rectangle (13,1.75);
  \node[anchor=east, font=\small\bfseries] at (-0.2,1.475) {Spk\,C};

  %% === Prediction point and outcome ===
  \draw[dashed, thick] (10,0.2) -- (10,4.55);
  \node[anchor=south east] at (9.9,4.0) {prediction point};
  \node[anchor=south, font=\small\bfseries] at (12.1,3.9) {HOLD};

  %% === Padding and gap arrows (below the rows) ===
  \draw[<->, >=Stealth, thick] (8.5,0.9) -- (10,0.9);
  \node[anchor=east] at (8.4,0.9) {padding};
  \draw[<->, >=Stealth, thick] (8.5,0.5) -- (11.2,0.5);
  \node[anchor=west] at (11.3,0.5) {gap};

  %% === Context label ===
  \node[gray!80] at (4,0.7) {context ($\leq$10\,s)};

  %% === Time axis ===
  \draw[->, >=Stealth, thick] (-0.2,0.2) -- (13.5,0.2) node[right] {$t$};

\end{tikzpicture}
\caption{A turn event at a mutual-silence gap hold.
The dashed line is the prediction point, 0.1\,s of padding after the previous
speaker's last word. Up to 10\,s of context precede it (shaded). The rest of the gap (at most 3\,s) and
the following speech (hatched) are withheld.}
  \vspace{-2ex}
\label{fig:timeline}
\end{figure}
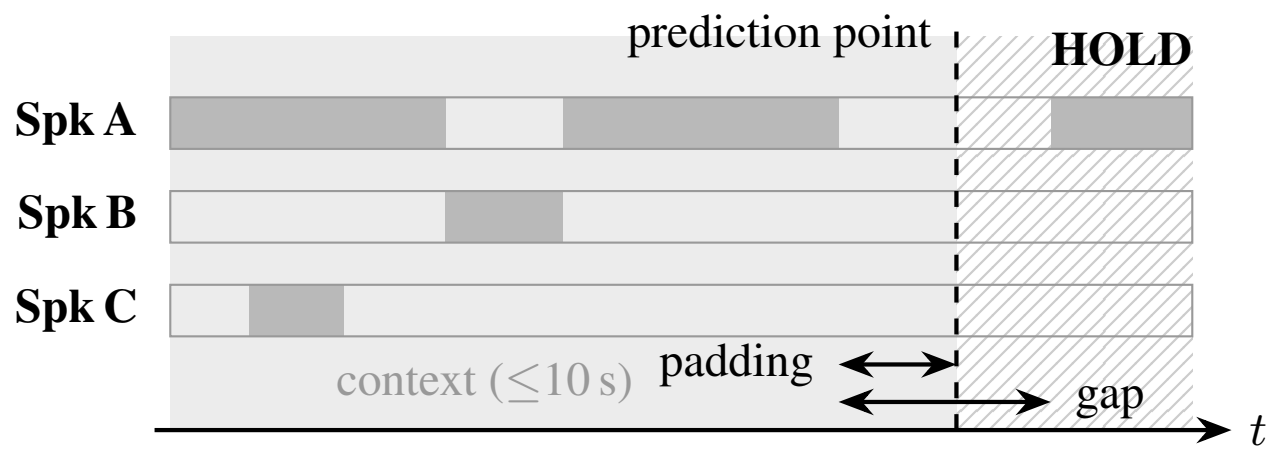

\textbf{Turn events} (Fig.~\ref{fig:timeline}). Words of the same speaker
separated by at most 0.1\,s are merged into inter-pausal units, which are
regrouped into utterances. A candidate event is a mutual-silence gap that
satisfies all of the following: the raw gap is shorter than 3.0\,s; after
0.1\,s of padding at least 0.1\,s of silence remains; the next speaker produces at least
1.0\,s of speech in the 2.0\,s after the gap and is the only active speaker
in the first 1.0\,s; and the gap contains no laughter or other vocal noise.
Holds outnumber shifts,
so holds were down-sampled at random to match, giving 865 shift and 865 hold
events on the test partition. 
% ===============================================================
\section{Human Benchmark and Evaluation}
\label{sec:study}

\subsection{Participants}
\label{sec:participants}

Participants were recruited through two channels: 85 people from the authors'
personal network and 103 participants on Prolific~\cite{prolific}. In total
188 people answered at least one of the 30 main-study trials.
Participation was voluntary and informed
consent was obtained before the task.  Each participant rated their own English
proficiency once on a five-level scale (native, near-native, advanced,
intermediate, beginner). The retained sample contains 82 native, 40
near-native and 32 advanced speakers.
\vspace{-1ex} 
\subsection{Task and Interface}
Each trial presents one clip: the audio and video from 10\,s before the
gap onset up to the gap onset. Face boxes labeled 0, 1, and 2 are drawn on
every frame from the corpus face tracks (Fig.~\ref{fig:interface}). Each trial
shows one clip of one event. The response buttons appear only after the clip
has played to the end once. Participants may replay the clip as often as they
wish, then pick the face they expect to speak next and rate their confidence
from 1 (``not at all confident'') to 5 (``very confident'').

Each participant judged 30 events in three rounds
of ten, each round containing five hold and five shift events, so that every
complete participant saw 15 events of each type. Two events from the same segment
less than 10\,s apart were never given to the same participant.

\vspace{-1ex} 
\subsection{Quality Filtering}
\label{sec:filter}

Participants are excluded by three rules, applied by the released script and
listed here with the number of participants each rule removes (rules
overlap). R1: self-reported proficiency below advanced (13). R2: automatic
flags: fewer than 4 saved trials, three or more skipped clips (18).
R3: Prolific submissions that were returned or timed out (6). Partial
sessions of at least 4 trials are kept, because per-participant accuracy is
a mean over that participant's own responses and does not require equal
counts. Of the 188 people, 34 are removed by at least one rule and 154 are
retained. Four of them took part in short follow-up runs on events with
tied votes; their responses are treated like every other response.

\subsection{Evaluation}
\label{sec:eval}

Human-Ind, RoleVAP, and MuVAP are measured on the same 1,730 test
events. Human-Maj is measured
on the events with an untied plurality (1,479 events under NSP scoring,
1,526 under hold/shift scoring). \emph{Hold/shift scoring} collapses a response to hold when the
chosen face is the previous speaker and to shift otherwise. RoleVAP predicts a current and a next role and cannot name the next
speaker from monaural audio, so it is scored on hold/shift only. For both
models the hold/shift decision is read out by a logistic regression fitted
on the model's embedding at the prediction point, using the AVCC training
partition (probe). MuVAP is also scored on NSP. It names the current
floor holder from its own output; on a probe decision of hold it predicts
that speaker, on shift the most probable other speaker. The hold/shift row
scores the probe decision alone, the NSP row the named speaker, which
also depends on identifying the floor holder. We report balanced
accuracy, the mean of hold and shift accuracy, overall and per speaker
count. Human-Ind is the mean over participants of their hold and shift
accuracy in a cell. Human-Maj assigns each event the label with the most votes
and drops ties.

% ===============================================================

\section{Results}
\label{sec:results}
\begin{table}[t]
  \caption{Accuracy (\%) on the 1,730 AVCC test events (865 hold / 865
  shift). Left: balanced accuracy by scoring
  and speaker count. Right: hold and shift accuracy by class.
  Human-Ind: mean per-participant accuracy ($N{=}154$). Human-Maj:
  plurality label per event, ties excluded (251 under NSP scoring,
  204 under hold/shift); the model rows beneath it are scored on the
  same retained events. Random Baseline: expected accuracy of a uniform
  guess over the faces (NSP) or over hold and shift, weighted by the events
  in each cell. Human-Maj baseline is 41.5.
  Scoring and the probe readout as in Sec.~\ref{sec:eval}.}
  \label{tab:results}
  \centering
  \scriptsize
  \renewcommand{\arraystretch}{1.12}
  \setlength{\aboverulesep}{0.25ex}
  \setlength{\belowrulesep}{0.35ex}
  \setlength{\tabcolsep}{2.2pt}
  \begin{tabular}{@{}l ccc ccc cc cc@{}}
    \toprule
    & \multicolumn{6}{c}{\textbf{Balanced accuracy}} & \multicolumn{4}{c}{\textbf{Per class}} \\
    \cmidrule(lr){2-7}\cmidrule(lr){8-11}
    & \multicolumn{3}{c}{NSP} & \multicolumn{3}{c}{Hold/shift} & \multicolumn{2}{c}{NSP} & \multicolumn{2}{c}{Hold/shift} \\
    \cmidrule(lr){2-4}\cmidrule(lr){5-7}\cmidrule(lr){8-9}\cmidrule(lr){10-11}
    \textbf{System} & 2-spk & 3-spk & All & 2-spk & 3-spk & All & hold & shift & hold & shift \\
    \midrule
    \textit{All events} & & & & & & & & & & \\
        Random Baseline              & 50.0 & 33.3 & 41.4 & 50.0 & 50.0 & 50.0 & 41.4 & 41.4 & 50.0 & 50.0 \\
    Human-Ind                    & 63.1 & 57.0 & 59.5 & 63.1 & 65.1 & 64.0 & 63.6 & 55.4 & 63.6 & 64.4 \\
    RoleVAP (probe)~\cite{muvap} & --   & --   & --   & 65.2 & 67.7 & 66.5 & --   & --   & 54.9 & 78.2 \\
    MuVAP (probe)~\cite{muvap}   & 67.5 & 59.4 & 63.3 & 66.7 & 69.0 & 67.9 & 62.8 & 63.8 & 58.7 & 77.0 \\

    \midrule
    \textit{Non-tied events} & & & & & & & & & & \\
    Human-Maj                    & 67.6 & 67.8 & 67.7 & 67.6 & 72.3 & 70.1 & 72.5 & 62.9 & 70.2 & 70.1 \\
    RoleVAP (probe)   & --   & --   & --   & 65.9 & 68.6 & 67.3 & --   & --   & 55.4 & 79.2 \\
    MuVAP (probe)     & 68.1 & 59.8 & 63.9 & 67.4 & 70.5 & 69.1 & 63.8 & 64.0 & 59.4 & 78.7 \\

    \bottomrule
  \end{tabular}
    \vspace{-1.5ex} 
\end{table}

Under NSP scoring in Table~\ref{tab:results}, the mean participant reaches 59.5\%, with 63.1\% on
two-speaker and 57.0\% on three-speaker events, a
loss of 6.1 points; the majority vote reaches 67.7\% on the
1,479 untied events. MuVAP reaches 63.3\%, 3.8 points above the mean
participant and 4.4 points below the majority vote, and loses 8.1 points
from two- to three-speaker events (67.5\% and 59.4\%). On two-speaker
events its NSP accuracy higher than its hold/shift accuracy (66.7\%)
because a misidentified floor holder and a wrong turn decision can cancel
and still name the right face.

Under hold/shift scoring where RoleVAP can be measured on, the
mean participant reaches 64.0\% and the majority vote 70.1\%
on 1,526 untied events. RoleVAP reaches 66.5\%, 2.5 points above the mean
participant and 3.6 points below the majority vote; MuVAP reaches 67.9\%,
1.4 points above the audio-only model. Both models gain with the number
of speakers under this scoring (65.2\% and 67.7\% for RoleVAP, 66.7\% and
69.0\% for MuVAP), as do the participants when they need not name the speaker (63.1\% and 65.1\%).

Participants find holds easier than shifts when they must name the speaker:
the majority vote reaches 72.5\% on holds and 62.9\% on shifts under
NSP scoring, while under hold/shift scoring the two are level. Both models show the opposite balance under hold/shift
scoring, RoleVAP 54.9\% on holds and 78.2\% on shifts, MuVAP 58.7\% and
77.0\%, so their lead over the mean participant comes from shifts that
people miss, and they lose on holds that people get right. Under NSP scoring
MuVAP's hold and shift accuracy are level (62.8\% and 63.8\%); on
three-speaker shifts it names the incoming speaker in 57.4\% of events.

\begin{figure}[t]
  \centering

  \includegraphics[width=\columnwidth]{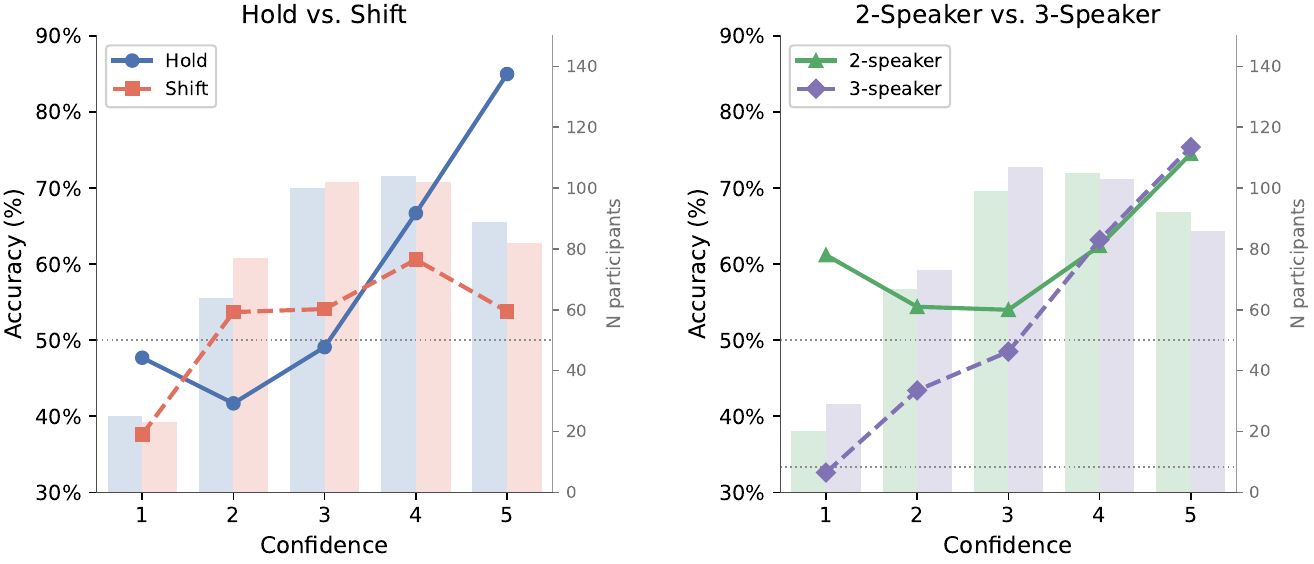}
  \caption{
  Per-participant accuracy by self-reported confidence. Left: hold (blue)
  vs.\ shift (orange). Right: two-speaker (green) vs.\ three-speaker
  (purple) clips. Bars give the number of participants contributing to each
  point; dotted lines mark chance (50\% and 33\%).}
  \label{fig:confidence}
  \vspace{-1.5ex} 
\end{figure}

Fig.~\ref{fig:confidence} splits per-participant accuracy by confidence.
Hold accuracy rises from 43.2\% at confidence 1 to
85.0\% at confidence 5, while shift accuracy
stays between 51.5\% and 60.1\% at ratings 2 to 5 and peaks at rating 4,
not 5. Speaker count
shows a different pattern: at confidence 1, three-speaker accuracy
(35.6\%) and two-speaker accuracy (52.8\%) both sit close to their chance
levels where at confidence 5 both exceed 75\%.

% ===============================================================
\vspace{-1ex} 
\section{Discussion}
\label{sec:discussion}

The human reference gives two anchors: either treating each judgement as a
datapoint (Human-Ind) or only keeping events where there was a plurality
(Human-Maj). Human-Maj is defined on fewer events, so the models are
compared to it on the same retained events (lower block of
Table~\ref{tab:results}); comparisons to Human-Ind use all events. On
balanced accuracy both models fall between the two anchors. On the
hold/shift decision RoleVAP is 2.5 points above the mean participant and
2.8 below the crowd; MuVAP is 3.9 above and 1.0 below. On NSP, MuVAP is 3.8
above the mean participant and 3.8 below the crowd. Adding video to audio
buys 1.4 points on hold/shift. The per-class columns show where the gap is:
both models predict shift too often, scoring above the crowd on shifts but
below the mean participant on holds. A current model already beats one
person watching the clip; the 1 to 4 points to the crowd, mostly on holds,
are the remaining headroom.

Going from two to three speakers costs participants 6.1 points on NSP
and nothing on hold/shift, where they gain 2.0. MuVAP behaves the same way,
8.1 points down on NSP and 2.3 up on hold/shift, and RoleVAP gains 2.5. The extra speaker does not make the change harder to detect shift or hold but
it makes the next speaker harder to pin point exactly who will take the turn. For MuVAP the floor holder is not the
problem: it is identified in 93.4\% of events, and on those events alone
NSP accuracy rises only from 63.3\% to 64.4\%. The loss is in choosing
between the two candidates, where MuVAP names the right one in 57.4\% of
three-speaker shifts.

Fig.~\ref{fig:confidence} shows that participants know when a hold is coming:
hold accuracy rises from 43.2\% to 85.0\% across the five ratings. For shifts
confidence carries almost no information, with accuracy between 51.5\% and
60.1\% and a peak at rating 4. At the lowest rating both two- and
three-speaker accuracy sit at chance, so low confidence is well calibrated.
The released ratings thus mark the events people themselves cannot decide,
and those are shifts.

Votes tie on 251 events under NSP scoring; 151 of them are shifts and 145 come from three-speaker videos, so a tie is most often a multiparty shift. Many of these are barge-ins or open questions put to the group: neither the audio nor the preceding context singles out a speaker, and the only cue is visual, the incoming speaker starting to move just before the gap.

\textbf{Limitations.} Participants were third-party observers of clips of at most 10 s, and they answered at the gap onset without seeing or hearing the gap itself. Any cue that appears inside the gap is therefore lost to them, including the fast visual cues an incoming speaker gives just before starting to talk. This makes short gaps easier than long ones: when the gap is short, the prediction point lies close to the next turn, so more of the next speaker's cues falls inside the visible context.
% ===============================================================
\vspace{-2ex} 
\section{Conclusion}
\label{sec:conclusion}

 We have extended AVCC, a corpus of two- and three-party conversation
recorded with a single camera and a single audio channel, together with a
speaker-disjoint test partition of 865 shift and 865 hold events, word-level
transcripts, and a crowdsourced human benchmark collected at the same
prediction points used to evaluate models. With a readout fitted on the
AVCC training partition, an audio-only Role-Relative VAP model scores
66.5\% on the hold/shift decision, against 64.0\% for the mean
participant and 70.1\% for the plurality vote, and the multimodal MuVAP
model reaches 67.9\%, and 63.3\% when it has to name the next
speaker. Scored on next-speaker identity,
participants drop 6.1 points from two to three speakers while their hold/shift
accuracy does not, so the third speaker makes it harder to name the incoming speaker, not to detect the change; MuVAP loses 8.1 points on the
same step and names the incoming speaker in 57.4\% of three-speaker shifts.
Participants predict holds better than shifts when naming the speaker, both models the reverse on the hold/shift decision, and participants' shift accuracy barely
increases with their confidence. In three-party shifts the next speaker has to be picked out of the
group. Neither audio alone nor a single observer does this reliably, and
that is where we expect the benchmark to be most useful.

\section{Acknowledgment}
\vspace{-1ex}
This work was supported by the Wallenberg AI, Autonomous Systems and Software
Program (WASP) funded by the Knut and Alice Wallenberg Foundation (KAW). The
computations and data handling were enabled by the Berzelius resource provided
by the Knut and Alice Wallenberg Foundation at the National Supercomputer
Centre. The authors have no relevant financial or nonfinancial interests to
disclose. Generative AI tools were used only to edit the text
for grammar; they generated no scientific content, figures, or code.

\section{Compliance with Ethical Standards}
\vspace{-1ex}
The human benchmark study followed the principles of the Declaration of
Helsinki. It involved no physical intervention, no method intended to affect
participants, no risk of harm, and no sensitive personal data, so under the
Swedish Ethics Review Act (2003:460) it did not require approval by the Swedish
Ethical Review Authority. Participants were informed of the purpose of the
study, gave informed consent before the task, could withdraw at any time, and
Prolific participants were paid at the platform's recommended rate. The corpus
consists of publicly available online videos.

% References should be produced using the bibtex program from suitable
% BiBTeX files (here: strings, refs, manuals). The IEEEbib.bst bibliography
% style file from IEEE produces unsorted bibliography list.
% -------------------------------------------------------------------------
\bibliographystyle{IEEEbib}
\bibliography{refs}

@article{Skantze2021,
  author  = {Skantze, Gabriel},
  title   = {Turn-taking in conversational systems and human-robot interaction: A review},
  journal = {Comput. Speech Lang.},
  volume  = {67},
  pages   = {101178},
  year    = {2021},
}

@inproceedings{vap,
  author    = {Ekstedt, Erik and Skantze, Gabriel},
  title     = {Voice Activity Projection: Self-supervised Learning of Turn-taking Events},
  booktitle = {Proc.\ Interspeech},
  pages     = {236--240},
  year      = {2022},
}

@article{muvap,
  author  = {Qi, Haotian and Skantze, Gabriel},
  title   = {{MuVAP}: Multimodal Multiparty Voice Activity Projection for Turn-taking
             Prediction in the Wild},
  journal = {arXiv preprint arXiv:2606.16731},
  year    = {2026},
}

@inproceedings{ekstedt2020turngpt,
  author    = {Ekstedt, Erik and Skantze, Gabriel},
  title     = {{TurnGPT}: A Transformer-based Language Model for Predicting Turn-taking
               in Spoken Dialog},
  booktitle = {Proc.\ EMNLP (Findings)},
  pages     = {2981--2990},
  year      = {2020},
}

@inproceedings{roddy2018,
  author    = {Roddy, Matthew and Skantze, Gabriel and Harte, Naomi},
  title     = {Investigating Speech Features for Continuous Turn-taking Prediction Using {LSTMs}},
  booktitle = {Proc.\ Interspeech},
  pages     = {1926--1930},
  year      = {2018},
}

@article{sacks:74,
  author  = {Sacks, Harvey and Schegloff, Emanuel A. and Jefferson, Gail},
  title   = {A Simplest Systematics for the Organization of Turn-taking for Conversation},
  journal = {Language},
  volume  = {50},
  number  = {4},
  pages   = {696--735},
  year    = {1974},
}

@article{kendon1967gaze,
  author  = {Kendon, Adam},
  title   = {Some Functions of Gaze Direction in Social Interaction},
  journal = {Acta Psychologica},
  volume  = {26},
  pages   = {22--63},
  year    = {1967},
}

@article{duncan1972some,
  author  = {Duncan, Starkey},
  title   = {Some Signals and Rules for Taking Speaking Turns in Conversations},
  journal = {J. Pers. Soc. Psychol.},
  volume  = {23},
  number  = {2},
  pages   = {283--292},
  year    = {1972},
}

@book{Duncan1977FacetofaceIR,
  author    = {Duncan, Starkey and Fiske, Donald W.},
  title     = {Face-to-Face Interaction: Research, Methods, and Theory},
  publisher = {Lawrence Erlbaum Associates},
  year      = {1977},
}

@article{stivers2009universals,
  author  = {Stivers, Tanya and others},
  title   = {Universals and Cultural Variation in Turn-taking in Conversation},
  journal = {Proc. Natl. Acad. Sci. USA},
  volume  = {106},
  number  = {26},
  pages   = {10587--10592},
  year    = {2009},
}

@article{levinson_torreira:15,
  author  = {Levinson, Stephen C. and Torreira, Francisco},
  title   = {Timing in Turn-taking and Its Implications for Processing Models of Language},
  journal = {Frontiers in Psychology},
  volume  = {6},
  pages   = {731},
  year    = {2015},
}

@article{de2006projecting,
  author  = {de Ruiter, Jan P. and Mitterer, Holger and Enfield, Nick J.},
  title   = {Projecting the End of a Speaker's Turn: A Cognitive Cornerstone of Conversation},
  journal = {Language},
  volume  = {82},
  number  = {3},
  pages   = {515--535},
  year    = {2006},
}

@article{gravano2011turn,
  author  = {Gravano, Agust{\'\i}n and Hirschberg, Julia},
  title   = {Turn-taking Cues in Task-oriented Dialogue},
  journal = {Comput. Speech Lang.},
  volume  = {25},
  number  = {3},
  pages   = {601--634},
  year    = {2011},
}

@inproceedings{godfrey1992switchboard,
  author    = {Godfrey, John J. and Holliman, Edward C. and McDaniel, Jane},
  title     = {{SWITCHBOARD}: Telephone Speech Corpus for Research and Development},
  booktitle = {Proc.\ ICASSP},
  pages     = {517--520},
  year      = {1992},
}

@inproceedings{cieri2004fisher,
  author    = {Cieri, Christopher and Miller, David and Walker, Kevin},
  title     = {The {Fisher} Corpus: A Resource for the Next Generations of Speech-to-Text},
  booktitle = {Proc.\ LREC},
  pages     = {69--71},
  year      = {2004},
}

@article{calhoun2010nxt,
  author  = {Calhoun, Sasha and others},
  title   = {The {NXT}-format {Switchboard} Corpus: A Rich Resource for Investigating
             the Syntax, Semantics, Pragmatics and Prosody of Dialogue},
  journal = {Lang. Resour. Eval.},
  volume  = {44},
  pages   = {387--419},
  year    = {2010},
}

@inproceedings{carletta2006ami,
  author    = {Carletta, Jean and others},
  title     = {The {AMI} Meeting Corpus: A Pre-announcement},
  booktitle = {Proc.\ MLMI},
  pages     = {28--39},
  year      = {2006},
}

@inproceedings{cafaro2017noxi,
  author    = {Cafaro, Angelo and others},
  title     = {The {NoXi} Database: Multimodal Recordings of Mediated Novice-Expert Interactions},
  booktitle = {Proc.\ ACM ICMI},
  pages     = {350--359},
  year      = {2017},
}

@inproceedings{roth2020ava,
  author    = {Roth, Joseph and others},
  title     = {{AVA-ActiveSpeaker}: An Audio-Visual Dataset for Active Speaker Detection},
  booktitle = {Proc.\ ICASSP},
  pages     = {4492--4496},
  year      = {2020},
}

@inproceedings{msdwild,
  author    = {Liu, Tao and others},
  title     = {{MSDWild}: Multi-modal Speaker Diarization Dataset in the Wild},
  booktitle = {Proc.\ Interspeech},
  pages     = {1476--1480},
  year      = {2022},
}

@article{reece2023candor,
  author  = {Reece, Andrew and others},
  title   = {The {CANDOR} Corpus: Insights from a Large Multimodal Dataset of
             Naturalistic Conversation},
  journal = {Science Advances},
  volume  = {9},
  number  = {13},
  pages   = {eadf3197},
  year    = {2023},
  doi     = {10.1126/sciadv.adf3197},
}

@article{northcutt2023egocoms,
  author  = {Northcutt, Curtis G. and others},
  title   = {{EgoCom}: A Multi-person Multi-modal Egocentric Communications Dataset},
  journal = {IEEE Trans. Pattern Anal. Mach. Intell.},
  volume  = {45},
  number  = {6},
  pages   = {6783--6793},
  year    = {2023},
}

@article{roxo2024wasd,
  author  = {Roxo, Tiago and Proen{\c{c}}a, Hugo},
  title   = {{WASD}: A Wilder Active Speaker Detection Dataset},
  journal = {IEEE Trans. Pattern Anal. Mach. Intell.},
  volume  = {46},
  number  = {12},
  pages   = {8431--8443},
  year    = {2024},
}

@inproceedings{mueller2021multimediates,
  author    = {M{\"u}ller, Philipp and others},
  title     = {{MultiMediate}: Multi-modal Group Behaviour Analysis for Artificial Mediation},
  booktitle = {Proc.\ ACM Multimedia},
  pages     = {4878--4882},
  year      = {2021},
}

@article{jiang2026turnbench,
  author  = {Jiang, Fei and others},
  title   = {{TurnBench}: A Multi-domain Benchmark for Turn-taking Dynamics in Spoken Dialogue},
  journal = {arXiv preprint arXiv:2608.25218},
  year    = {2026},
}

@inproceedings{elmers2025triadic,
  author    = {Elmers, Mikey and others},
  title     = {Triadic Multi-party Voice Activity Projection for Turn-taking in Spoken
               Dialogue Systems},
  booktitle = {Proc.\ Interspeech},
  pages     = {3015--3019},
  year      = {2025},
}

@inproceedings{heo2025gaze,
  author    = {Heo, Seonguk and others},
  title     = {Gaze-enhanced Multimodal Turn-taking Prediction in Triadic Conversations},
  booktitle = {Proc.\ Interspeech},
  pages     = {1068--1072},
  year      = {2025},
}

@article{voxstral,
  author  = {Liu, Alexander H. and others},
  title   = {Voxtral},
  journal = {arXiv preprint arXiv:2507.13264},
  year    = {2025},
}

@inproceedings{mcauliffe2017montreal,
  author    = {McAuliffe, Michael and others},
  title     = {Montreal Forced Aligner: Trainable Text-Speech Alignment Using {Kaldi}},
  booktitle = {Proc.\ Interspeech},
  pages     = {498--502},
  year      = {2017},
}

@inproceedings{guo2022scrfd,
  author    = {Guo, Jia and others},
  title     = {Sample and Computation Redistribution for Efficient Face Detection},
  booktitle = {Proc.\ ICLR},
  year      = {2022},
}

@inproceedings{deng2019arcface,
  author    = {Deng, Jiankang and others},
  title     = {{ArcFace}: Additive Angular Margin Loss for Deep Face Recognition},
  booktitle = {Proc.\ CVPR},
  pages     = {4690--4699},
  year      = {2019},
}

@inproceedings{talknet,
  author    = {Tao, Ruijie and others},
  title     = {Is Someone Speaking? Exploring Long-term Temporal Features for
               Audio-Visual Active Speaker Detection},
  booktitle = {Proc.\ ACM Multimedia},
  pages     = {3927--3935},
  year      = {2021},
}

@article{kuhn1955hungarian,
  author  = {Kuhn, Harold W.},
  title   = {The {Hungarian} Method for the Assignment Problem},
  journal = {Nav. Res. Logist. Q.},
  volume  = {2},
  number  = {1--2},
  pages   = {83--97},
  year    = {1955},
}

@article{ishii2016respiration,
  author  = {Ishii, Ryo and Otsuka, Kazuhiro and Kumano, Shiro and Yamato, Junji},
  title   = {Using Respiration to Predict Who Will Speak Next and When in Multiparty Meetings},
  journal = {ACM Trans. Interact. Intell. Syst.},
  volume  = {6},
  number  = {2},
  pages   = {20:1--20:20},
  year    = {2016},
}

@inproceedings{ishii2016mouth,
  author    = {Ishii, Ryo and Kumano, Shiro and Otsuka, Kazuhiro},
  title     = {Analyzing Mouth-opening Transition Pattern for Predicting Next Speaker
               in Multi-party Meetings},
  booktitle = {Proc.\ ICMI},
  pages     = {209--216},
  year      = {2016},
}

@article{jokinen2013gaze,
  author  = {Jokinen, Kristiina and Furukawa, Hirohisa and Nishida, Masafumi and Yamamoto, Seiichi},
  title   = {Gaze and Turn-taking Behavior in Casual Conversational Interactions},
  journal = {ACM Trans. Interact. Intell. Syst.},
  volume  = {3},
  number  = {2},
  pages   = {12:1--12:30},
  year    = {2013},
}

@inproceedings{malik2020speaks,
  author    = {Malik, Usman and Saunier, Julien and Funakoshi, Kotaro and Pauchet, Alexandre},
  title     = {Who Speaks Next? {Turn} Change and Next Speaker Prediction in Multimodal
               Multiparty Interaction},
  booktitle = {Proc.\ IEEE ICTAI},
  pages     = {349--354},
  year      = {2020},
}

@inproceedings{bohus2011decisions,
  author    = {Bohus, Dan and Horvitz, Eric},
  title     = {Decisions About Turns in Multiparty Conversation: From Perception to Action},
  booktitle = {Proc.\ ICMI},
  pages     = {153--160},
  year      = {2011},
}

@misc{prolific,
  author       = {{Prolific Academic Ltd.}},
  title        = {Prolific: A Platform for Recruiting Online Research Participants},
  howpublished = {\url{https://www.prolific.com}},
  year         = {2014},
}

\end{document}